%% file: egpaper_for_review.tex
\documentclass[10pt,twocolumn,letterpaper]{article}

\usepackage{cvpr}
\usepackage{times}
\usepackage{epsfig}
\usepackage{graphicx}
\usepackage{amsmath}
\usepackage{amssymb}
\usepackage{booktabs}       % professional-quality tables
\usepackage{multirow}
\usepackage[subrefformat=parens]{subcaption}
\usepackage{adjustbox}
\usepackage{balance}

\def\sub#1{_{\rm #1}}

\def\eg{{\it e.g.}}

\def\ie{{\it i.e.}}

\makeatletter
\newcommand{\printfnsymbol}[1]{%
  \textsuperscript{\@fnsymbol{#1}}%
}
\makeatother

\def\arraystrechlen{.88}

\usepackage[pagebackref=true,breaklinks=true,letterpaper=true,colorlinks,bookmarks=false]{hyperref}

\cvprfinalcopy % *** Uncomment this line for the final submission

\def\cvprPaperID{2294} % *** Enter the CVPR Paper ID here

\ifcvprfinal\fi
\begin{document}

%%%%%%%%% TITLE
\title{Crowd Density Forecasting by Modeling Patch-based Dynamics}

\author{Hiroaki Minoura\thanks{Work done as an intern at OMRON SINIC X}\;\thanks{Equal contribution}\\
Chubu University\\
Aichi, Japan\\
{\tt\small himi1208@mprg.cs.chubu.ac.jp}
% For a paper whose authors are all at the same institution,
% omit the following lines up until the closing ``}''.
% Additional authors and addresses can be added with ``\and'',
% just like the second author.
% To save space, use either the email address or home page, not both
\and
Ryo Yonetani\printfnsymbol{2}\;\;\;Mai Nishimura\;\;\;Yoshitaka Ushiku\\
OMRON SINIC X\\
Tokyo, Japan\\
{\tt\small \{ryo.yonetani,mai.nishimura,yoshitaka.ushiku\}@sinicx.com}
}

\maketitle
%\thispagestyle{empty}

%%%%%%%%% ABSTRACT
\begin{abstract}
Forecasting human activities observed in videos is a long-standing challenge in computer vision, which leads to various real-world applications such as mobile robots, autonomous driving, and assistive systems. In this work, we present a new visual forecasting task called crowd density forecasting. Given a video of a crowd captured by a surveillance camera, our goal is to predict how that crowd will move in future frames. To address this task, we have developed the patch-based density forecasting network (PDFN), which enables forecasting over a sequence of crowd density maps describing how crowded each location is in each video frame. PDFN represents a crowd density map based on spatially overlapping patches and learns density dynamics patch-wise in a compact latent space. This enables us to model diverse and complex crowd density dynamics efficiently, even when the input video involves a variable number of crowds that each move independently. Experimental results with several public datasets demonstrate the effectiveness of our approach compared with state-of-the-art forecasting methods.
\end{abstract}

%%%%%%%%% BODY TEXT
\section{Introduction}
\label{sec:intro}

We envision a future intelligent system that monitors and supports safe transport in the city. Specifically, imagine a surveillance camera system mounted on the corner of a street. If the system is able to visually forecast how crowded the location in its field of view will be in the near future, that is valuable information for people or vehicles nearby to avoid potential collisions with the surrounding crowd. Moreover, once many surveillance systems across the world gain such a visual forecasting ability, large-scale prediction of future crowded events will enhance a variety of applications including city-scale traffic analysis \cite{Zhang_2017_AAAI,Zonoozi_2018_IJCAI} and city attribute modeling \cite{zhou2014recognizing}. Toward this goal, we propose a new visual forecasting challenge called \emph{crowd density forecasting} that predicts how a crowd will move in a video.

\begin{figure}[t]
  \begin{center}
    \includegraphics[width=\linewidth]{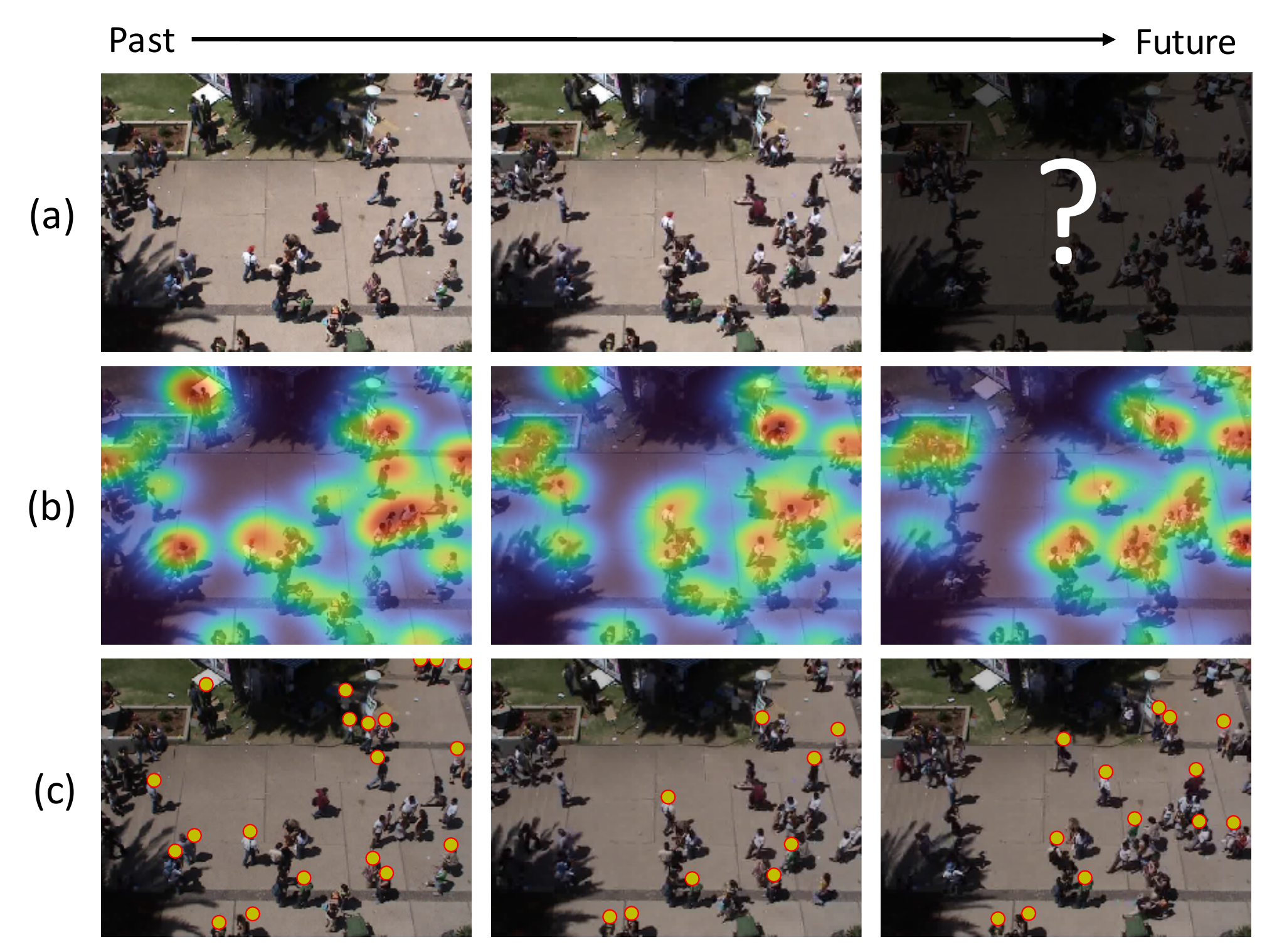}
    \caption{\textbf{Crowd density forecasting.} Given (a) a sequence of video frames of a crowd, we forecast how that crowd will move in unseen future frames in the form of (b) crowd density maps (maps of how crowded each location is) rather than (c) detecting, tracking, and forecasting future locations of each individual (shown in yellow circles).} 
    \label{fig:teaser}
  \end{center}
\end{figure}

To achieve this new task, one relevant technique studied extensively in the computer vision community is trajectory forecasting, which aims to predict the future trajectories of moving targets (\eg, pedestrians, vehicles) from their locomotion history. Some recent studies have indeed focused on forecasting the trajectories of groups of people \cite{Alahi_2016_CVPR,Gupta_2018_CVPR,Ivanovic_2019_ICCV,sadeghian2019sophie,vemula2018social,Xu_2018_CVPR}, and trajectory forecasting has been recognized as a fundamental technique to empower various intelligent systems such as mobile robots \cite{Bera_2017_IROS,Kretzschmar_2014_ICRA,Luber_2010_ICRA,Trautman_2015_IJRR}, autonomous driving \cite{altche2017lstm,Chandra_2019_CVPR,houenou2013vehicle,Jain_2019_CoRL,Lee_2017_CVPR,rhinehart2019precog}, and blind navigation \cite{manglik2019future}. However, most works have assumed each individual trajectory as both input and output and necessitated accurate detection and tracking of targets. We argue that this requirement prevents us from applying existing trajectory forecasting methods to videos of a crowd, since pedestrian detection and tracking can easily fail to work in congested areas, such as the one depicted in Fig.~\ref{fig:teaser} (c).

Based on this insight, we propose to forecast a map of how crowded each location will be in future frames (\ie, future crowd density maps) directly, rather than tracking each individual trajectory. As shown in Fig.~\ref{fig:teaser} (b), we train a forecasting model over a sequence of crowd density maps extracted from past to future frames. Doing so will allow the model to achieve the dynamics of a crowd motion while bypassing the use of pedestrian detection and tracking.

The main technical difficulty with the proposed approach is how to model crowd density maps to forecast them efficiently. In particular when input videos are captured by a camera covering wide areas, they often include multiple groups of people that each move independently. Moreover, the number of observed groups and the entire crowdedness of a scene can vary extensively depending on the pose and position of the cameras as well as on the types of scene (\eg, a busy shopping mall or a quiet street corner). This makes the spatiotemporal dynamics of crowd density maps diverse and complex, and ultimately difficult to forecast.

\begin{figure*}[t]
  \begin{center}
    \includegraphics[width=\linewidth]{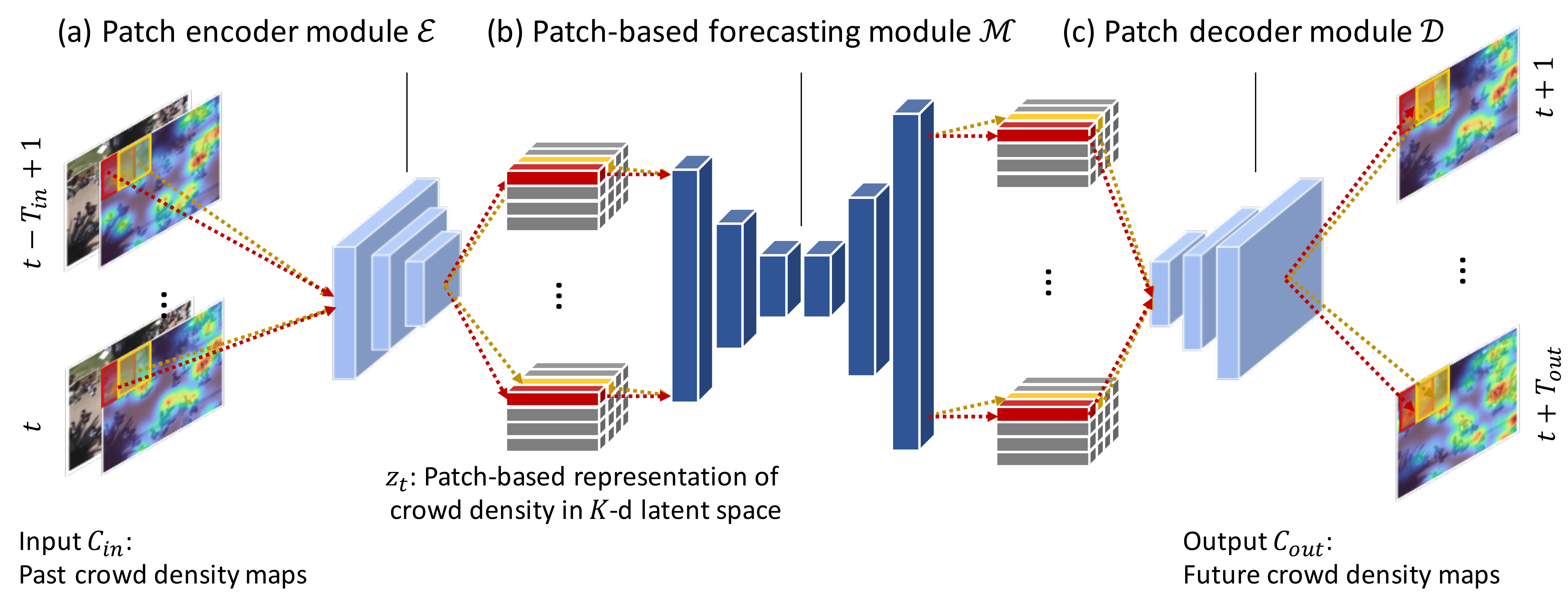}
    \caption{\textbf{Patch-based density forecasting network (PDFN).} Given a history of crowd density maps, PDFN a) decomposes the maps into spatially overlapping patches and learns their latent representation, and b) learns to forecast patch-based density dynamics in the latent space. c) The forecasted results are decoded to reconstruct a sequence of future crowd density maps.}
    \label{fig:overview}
  \end{center}
\end{figure*}

To address this difficulty, we develop a new forecasting model called the \emph{patch-based density forecasting network (PDFN)}. PDFN better models diverse and complex crowd dynamics consisting of various numbers of groups of people by decomposing crowd density maps into a collection of spatially overlapping patches, and learning patch-based density dynamics that is simpler and less diverse (see also Figure~\ref{fig:overview}). To cope with the discontinuities of forecasted results between adjacent patches, PDFN learns the representations of crowd density patches with a fully convolutional encoder-decoder architecture and forecasts patch-based dynamics in a latent feature space. In this way, PDFN can reconstruct future crowd density maps with fewer discontinuities by decoding predicted feature vectors.

We evaluate PDFN on several public datasets including FDST~\cite{fang2019locality}, a well-annotated video dataset developed initially for crowd counting tasks, as well as the UCY dataset that has been commonly used for a trajectory forecasting benchmark~\cite{lerner2007crowds}. We confirm that our approach is able to forecast future crowd density accurately, even when baseline methods with state-of-the-art trajectory forecasting~\cite{Alahi_2016_CVPR,Ivanovic_2019_ICCV} perform poorly due to unstable detection and tracking under congested conditions.

\paragraph{Our Contributions.} 1) We present crowd density forecasting, a new task that requires the prediction of future crowd density maps of a scene given its history; 2) We develop a patch-based density forecasting network tailored to learn diverse and complex crowd density dynamics, which is confirmed to work well on multiple public datasets.

\section{Related Work}
\label{sec:related}

The task of crowd density forecasting can be encompassed by visual forecasting that has been recognized as a long-standing challenge in computer vision. Below, we briefly review some recent work mainly presented recently.

\paragraph{Trajectory Forecasting in Computer Vision.}
Visual forecasting of human activities has widely been studied but remains a challenging problem in computer vision. In particular, much work has been done recently on trajectory forecasting, \ie, predicting a trajectory of how individual targets will move, with the rapid development of temporal modeling based on deep recurrent networks and imitation learning (\eg, \cite{Kitani2012}; see \cite{rudenko2019human} for a recent extensive survey). One key technical challenge, which is also closely related to our work, is how to forecast the trajectories of pedestrians interacting with each other in a group. Major approaches in this vein include social force modeling~\cite{Dirk1995,mehran2009abnormal}, multi-agent frameworks~\cite{le2017coordinated,ma2017forecasting,rhinehart2019precog,Zhao_2019_CVPR}, pooling states among multiple recurrent models~\cite{Alahi_2016_CVPR,Choi_2019_ICCV,Gupta_2018_CVPR,Lee_2017_CVPR,sadeghian2019sophie,Xu_2018_CVPR,Zhang_2019_CVPR}, and interaction modeling via graph neural networks~\cite{Ivanovic_2019_ICCV,vemula2018social}. However, all these works assume that accurate pedestrian trajectories are given both in training and testing. This is not always a realistic assumption when one wants to forecast the behaviors of a crowd because a large number of people would appear very small in each video frame and come with significant occlusions, making it difficult to detect and track them reliably. Our work resolves this issue by taking sequences of crowd density maps as inputs and outputs instead.

\paragraph{Other Relevant Forecasting Tasks.} 
There is also some work tackling a similar forecasting task in other research domains such as robotics~\cite{Bera_2017_IROS,Kretzschmar_2014_ICRA,Luber_2010_ICRA,Trautman_2015_IJRR}, intelligent vehicles~\cite{altche2017lstm,Chandra_2019_CVPR,houenou2013vehicle,Jain_2019_CoRL,Lee_2017_CVPR,Thiede_2019_ICCV,rhinehart2018r2p2,rhinehart2019precog}, and geo-spatial data analysis~\cite{zhou2014recognizing}. For example, recent work proposed a method that took auto-camera videos and LiDAR data as input to forecast vehicle trajectories~\cite{Chang_2019_CVPR,luo2018fast}. While such additional inputs would improve the crowd density forecasting ability of our approach, we limit our study to the use of a single surveillance video. \cite{Zhang_2017_AAAI,Zonoozi_2018_IJCAI} have attempted to enable city-scale forecasting of vehicle motion by making use of GPS histories. Similar to our work, they learn crowd dynamics with a convolutional neural network. However, their models focus only on a specific scenario in which training and testing videos are captured in the same scene. In contrast, our approach can work on a variety of complex crowd density dynamics due to its novel patch-based representation.

\paragraph{Crowd Density Estimation.}
Lastly, our work is also relevant to crowd density estimation, which is another popular computer vision task aiming to detect and count crowds in still images or videos.  Despite a huge amount of prior work (see \cite{kang2018beyond,sindagi2018survey,zitouni2016advances} for extensive surveys), attempts on video-based crowd density estimation are relatively limited \cite{fang2019locality,liu2019geometric,xingjian2015convolutional,zhang2017fcn}, and their main focus lies in how to leverage temporal continuities for improving density estimation, rather than how to forecast future crowdedness.

\section{Crowd Density Forecasting}
\label{sec:proposed}

In this section, we first formulate the problem of crowd density forecasting (Section~\ref{subsec:problem}) and then present the proposed \emph{patch-based density forecasting network (PDFN)}. As illustrated in Figure~\ref{fig:overview}, PDFN consists of a) patch encoder and c) decoder modules (Section~\ref{subsec:ae}) as well as b) patch-based forecasting module (Section~\ref{subsec:tconv}) to better model diverse and complex crowd dynamics.

\subsection{Problem Setting}
\label{subsec:problem}

Consider a collection of video clips taken by fixed surveillance cameras installed in a variety of places, which each involve various numbers of groups of people. Given the first few frames of the video as input, our goal is to forecast how observed crowds will move in the subsequent frames. To this end, our work involves a pre-processing step that utilizes an off-the-shelf crowd density estimator (\eg, \cite{gao2019c}) for all the videos to obtain how crowded each location of each frame is in the form of a crowd density map.

Formally, let $c_t\in[0, 1]^{W\times H}$ be a crowd density map extracted from the $t$-th input video frame of size $(W, H)$. Then, we denote an input and an output sequence of crowd density maps of length $T\sub{in}$ and $T\sub{out}$ as $C\sub{in}=[c_{t-T\sub{in} + 1}, \dots, c_{t}]$ and $C\sub{out}=[c_{t+1}, \dots, c_{t+T\sub{out}}]$, respectively. In the following, we present a deep neural network that can predict $C\sub{out}$ from $C\sub{in}$.

\subsection{Patch Encoder and Decoder Modules}
\label{subsec:ae}

As shown in Fig.~\ref{fig:overview} (a), the patch encoder module first decomposes each crowd density map in $C\sub{in}$ into multiple patches spatially overlapping with each other, while simultaneously learning their compact feature representation. The reasoning behind this approach is that, even when input videos involve many groups of people, the spatial layout and dynamics of each group would be less diverse (\eg, they would follow f-formation~\cite{Kendon1990} when they are standing still to interact with each other). Accordingly, we expect that the crowd density maps extracted from such videos can be described by a combination of smaller and simpler patterns.

Specifically, we achieve patch-wise compact representations of crowd density maps by learning an auto-encoder. Let us denote by $z_t \in \mathbb{R}^{W'\times H'\times K}$, a $K$-dimensional feature map of size $(W', H')$ ($W'<W,\;H'<H$), which we obtain by feeding input density map $c_t$ to a fully convolutional encoder, \ie, $z_t = \mathcal{E}(c_t)$. We also introduce a fully convolutional patch decoder that projects $z_t$ back to the input space, \ie, $c'_t = \mathcal{D}(z_t)\in[0,1]^{W \times H}$. As depicted in Fig.~\ref{fig:overview} (a), the forward operation of $\mathcal{E}(c_t)$ can be viewed as encoding multiple spatially-overlapping patches into a $K$-dimensional latent space with a shared auto-encoder, thus enabling us to learn simpler spatiotemporal patterns of crowd density maps observed in a smaller region.

The patch encoder $\mathcal{E}$ and patch decoder $\mathcal{D}$ are trained jointly with a collection of crowd density maps. Given a mini-batch of input sequences of size $B$, which is represented by $\mathbf{C}=\{C\sub{in}^{(1)},\dots,C\sub{in}^{(B)}\}$, we minimize the following binary cross-entropy (BCE) loss $\mathcal{L}(\mathbf{C})$ with respect to the trainable parameters in $\mathcal{E}$ and $\mathcal{D}$:
\begin{equation}
    \mathcal{L}(\mathbf{C}) = \frac{1}{B}\sum_{b=1}^B\left(\frac{1}{T\sub{in}}\sum_{c\in C\sub{in}^{(b)}} \textit{BCE}(c, \mathcal{D}(\mathcal{E}(c)))\right).
\end{equation}

\subsection{Patch-based Forecasting Module}
\label{subsec:tconv}
The patch encoder $\mathcal{E}$ presented above makes it possible to learn patch-based crowd density dynamics in a compact $K$-dimensional latent feature space. Learning a dynamics model in this way has also been done in other literature, for example, with model-based reinforcement learning~\cite{ha2018recurrent} and imitation learning~\cite{zeng2017visual}. However, these works encode a whole video frame into a single feature vector, which is not necessarily effective to our problem setting because, as discussed in Section \ref{sec:intro}, our input videos would involve multiple groups of people, and therefore, a whole crowd density dynamics would become combinatorially diverse. Moreover, forecasting in the latent feature space is advantageous to cope with patch-wise outputs that are independent with each other. Because the patch decoder $\mathcal{D}$ takes the form of fully-deconvolutional networks, it can take into account spatially adjacent outputs when projecting them back to the input space, while encouraging reconstructed crowd density maps to be less discontinuous.

Specifically, consider sequences of the $K$-dimensional feature maps, which were obtained by applying the pre-trained patch encoder $\mathcal{E}$ to $C\sub{in}$ and $C\sub{out}$:
\begin{eqnarray}
    Z\sub{in} = \left[\mathcal{E}(c_{t-T\sub{in}+1}),\dots,\mathcal{E}(c_{t})\right]\in\mathbb{R}^{W'\times H' \times K\times T\sub{in}},\\
    Z\sub{out} = \left[\mathcal{E}(c_{t+1}),\dots,\mathcal{E}(c_{t+T\sub{out}})\right]\in\mathbb{R}^{W'\times H' \times K\times T\sub{out}}.
\end{eqnarray}
Then, we model a patch-based forecasting module $\mathcal{M}$ by a temporal convolution-deconvolution network that maps $Z\sub{in}$ to $Z\sub{out}$, \ie, $Z\sub{out}'=\mathcal{M}(Z\sub{in}) \in\mathbb{R}^{W'\times H' \times K\times T\sub{out}}$, as shown in Fig.~\ref{fig:overview} (b). Importantly, these temporal convolutions and deconvolutions are done for each location of $(W', H')$ independently, meaning that patch-based dynamics can be learned by a smaller shared network. Given a mini-batch of $K$-dimensional feature map sequences $\mathbf{Z}=\left\{(Z\sub{in}^{(1)}, Z\sub{out}^{(1)}),\dots,(Z\sub{in}^{(B)}, Z\sub{out}^{(B)})\right\}$, we learn $\mathcal{M}$ by minimizing the following mean-squared error (MSE):
\begin{equation}
    \mathcal{L}(\mathbf{Z}) = \frac{1}{B}\sum_{b=1}^B \left( \textit{MSE}(Z^{(b)}\sub{out},\mathcal{M}(Z^{(b)}\sub{in})) \right).
    \label{eq:mse}
\end{equation}

\section{Experiments}
In this section, we compare the proposed PDFN with several state-of-the-art forecasting methods under various conditions on several public datasets. 

\subsection{Datasets}
\label{subsec:datasets}
In our experiments, we employed the following datasets comprising a variety of indoor and outdoor scenes.
\begin{itemize}
    \item \textbf{FDST} \cite{fang2019locality} is a dataset originally developed for video crowd counting tasks, which contains 100 videos of a crowd captured in 15 diverse places\footnote{The FDST dataset available online at the time of our submission had ten distinct scenes divided into five videos and five other scenes divided into ten videos (so 100 videos in total), although the original version was reported to contain 13 scenes in~\cite{fang2019locality}.} with unique camera poses and positions. Each video comprises 150 frames at 30 fps, and the locations of pedestrians are fully annotated for each frame.
    \item \textbf{UCY (Crowds-by-Example Dataset)} \cite{lerner2007crowds} is a common dataset for trajectory forecasting. Following a standard benchmark scheme, we used `Zara 1' (9,031 frames), `Zara 2' (10,519 frames), and `University' (5,405 frames; all recorded at 25 fps), which each involved distinct scenes of a crowd captured by a similar oblique point of view. Unlike the FDST dataset, this dataset comprised annotations of pedestrians only sparsely at every ten frames. 
\end{itemize}

\subsection{Training Setup}
\label{subsec:setup}

\paragraph{Obtaining Crowd Density Maps.}
One critical design choice for PDFN is how to compute crowd density maps from videos. In this experiment, we adopted and compared two different variants. 1) \textbf{PDFN-C} that utilized a state-of-the-art crowd density estimator~\cite{wang2019learning} implemented in the C-3 framework~\cite{gao2019c} to estimate the degree of crowdedness per pixel directly from input video frames. We also tested 2) \textbf{PDFN-D} that took inputs from a state-of-the-art object detector called the feature pyramid networks (FPN)~\cite{lin2017feature} with a ResNet backbone~\cite{he2016deep} implemented in ChainerCV~\cite{ChainerCV2017} to detect pedestrians. For PDFN-D, we mapped the top-center location or each detected bounding box to the input image space to form the input crowd density maps that are compatible with what was obtained by the crowd density estimator. In this way, PDFN-D is able to forecast future crow density maps without tracking each detected individual. Although we performed the crowd density estimation and the pedestrian detection on input video frames sized $640\times 480$, the obtained crowd density maps were then downsized to $80\times80$ so that training and testing of PDFN were done in a reasonable time.

\input{result_table_fdst.tex}

\paragraph{Creating Ground-Truth Maps.}
For each dataset, we built one ground-truth crowd density map per frame. Similar to how pedestrian detection results were processed above, we first projected all the annotated locations of pedestrians onto the image space and then resized them properly to align with the size of network outputs\footnote{We observed that the head locations of each person were annotated in FDST and Zara 2, while Zara 1 and University sequences contained annotations of feet. To resolve this annotation inconsistency, we used our pedestrian detection results to train a support vector regressor mapping from vertical locations of people to their height, and modified foot annotations in Zara 1 and University sequences to point to heads.}. We emphasize here that these ground-truth crowd density maps were not used for training PDFN and other baseline methods we will introduce shortly. This makes our experimental setting more realistic because it is not always feasible to annotate a large number of people for every frame to train crowd density forecasting models.

\paragraph{Splitting Datasets.} 
From the sequences of crowd density maps, we created many training and testing samples consisting of 20 consecutive frames at 6 fps for FDST and at 5 fps for UCY by applying a temporal sliding window to the sequences. The first eight frames were then used as input and the remaining twelve frames as output (\ie, $T\sub{in}=8,\; T\sub{out}=12$), following a standard evaluation scheme for trajectory forecasting. For UCY, we conducted a leave-one-video-out cross validation across the three videos, while adding the `Hotel' sequence (19,350 frames captured at 25 fps) from the ETH dataset~\cite{pellegrini2009you} to all the training splits to increase the diversity of trained crowd density dynamics\footnote{Although the ETH dataset is also used popularly for trajectory forecasting benchmark, our preliminary experiments have revealed that it was difficult to employ it in our crowd density forecasting tasks because of its incomplete ground-truth annotations.}. For FDST, we split 100 videos into 60 training and 40 test subsets, which respectively contain 10 and 5 distinct scenes\footnote{We found the original split of FDST made 60 training and 40 testing videos with distinct behaviors of crowds but under the same 15 sets of scenes. In order to investigate how trained models were able to be generalized to unseen scenes, we created a new data split with the same numbers of training and test videos but without any scene overlap.}.

\input{result_table_figure_ucy.tex}

\subsection{Evaluation Schemes}
\paragraph{Smoothing Maps.}
Our performance evaluation involves a comparison between forecasted and ground-truth crowd density maps. To do this, we first smoothed both the input/output and the ground-truth crowd density maps with a spatiotemporal Gaussian filter of a pre-defined kernel size $\sigma$. This kernel size was used in our experiments to control how strictly we measured the difference between predicted and ground-truth maps, similar to the threshold given to IoU scores in object detection tasks \cite{lin2014microsoft}; if we set $\sigma$ small, we expect our forecasting results to fit the ground-truth more strictly. In contrast, we could allow methods to make a more approximated forecasting by setting a larger $\sigma$. Unless specified otherwise, we set $\sigma=3$ through the experiments. 

\paragraph{Evaluation Metrics}
 We followed the trajectory forecasting metrics that measured the average performance over $T\sub{out}$ frames and at the final ($t + T\sub{out}$-th) frame for each sample. The predicted map at the $\tau$-th frame, $c_\tau$, was evaluated against the corresponding ground-truth map $g_\tau$ based on the following Kullback-Leibler (KL) divergence $D_{\mathrm{KL}}$ and its inverse version $D_{\mathrm{IKL}}$.
\begin{equation}
D\sub{KL}(g_\tau || c_\tau)  =  \frac{1}{W\cdot H}\sum_{i, j} \bar{g}_\tau(i, j)\log\left(\frac{\bar{g}_\tau(i, j)}{\bar{c}_\tau(i, j)}\right),
\label{eq:kl}
\end{equation}
\begin{equation}
D\sub{IKL}(g_\tau || c_\tau)  = D\sub{KL}(c_\tau || g_\tau),
\label{eq:ikl}
\end{equation}
where $\bar{c}_\tau = c_\tau / \sum_{i,j} c_\tau(i,j)$,  $\bar{g}_\tau = g_\tau / \sum_{i,j} g_\tau(i,j)$ are the predicted and ground-truth maps normalized to be a probabilistic distribution, respectively, and $i, j$ are the indices given to each of the locations of maps. Intuitively, $D\sub{KL}$ corresponds to the recall metric measuring how much of the presence of future crowd density was predicted successfully by $c_\tau$. In contrast, $D\sub{IKL}$ measures how accurate $c_\tau$ is when it is high, corresponding to the precision metric. Moreover, we report the Jensen-Shannon (JS) divergence defined as follows, which is a symmetric divergence intuitively indicating the balance between precision and recall performances in our problem setting:
\begin{equation}
D\sub{JS}(g_\tau || c_\tau)  = \frac{1}{2}(D\sub{KL}(g_\tau || \frac{g_\tau + c_\tau}{2}) + D\sub{KL}(c_\tau || \frac{g_\tau + c_\tau}{2})).
\label{eq:jsd}
\end{equation}
Note that all these divergences are a non-negative metric, where lower scores mean better performances.

 \input{result_figure.tex}

\subsection{Implementation Details}
\label{subsec:implementation}

The PDFN-C and PDFN-D were implemented and trained as follows. All the implementations were done in the PyTorch framework \cite{paszke2017automatic}.
\begin{itemize}
    \item  \textbf{Patch Encoder and Decoder Modules.} We implemented a standard auto-encoder that had three convolution and three deconvolution layers, each followed by the rectifier linear unit (ReLU) activation~\cite{nair2010rectified} and respectively halve and double the spatial size of input data. With this architecture, $W'$ and $H'$ were set to $W'=H'=10$, and the dimension of latent feature vectors was empirically set to $K=16$. To encourage the module to learn representations of both crowded and less-crowded regions, we fed $\sqrt{c_t}$ instead of $c_t$.
    \item \textbf{Patch-based Forecasting Module.} We implemented a set of three temporal convolution layers and another set of three deconvolution layers with ReLU activation. Temporal strides were set to accept input and output sequences of length $T\sub{in}$ and $T\sub{out}$, respectively.
\end{itemize}
Moreover, to determine the effectiveness of the patch-based modeling utilized by our approach, we also implemented a degraded variant of PDFN, referred to as \textbf{DFN-C} and \textbf{DFN-D}, which encoded crowd density maps (created in the same way as those of PDFN by crowd density estimation or pedestrian detection) into a single feature vector and learned its dynamics with a temporal convolution-deconvolution model. Because DFN had to deal with more complex and diverse spatiotemporal patterns with a single feature vector, we set its dimension to $K=128$.

All of these modules were trained by the Adam optimizer~\cite{kingma2014adam} with mini-batches of $B=16$, where the number of iterations and learning rate were set to (1k, 0.001) for FDST and (100k, 0.005) for the UCY dataset.

\subsection{Baseline Methods}
Due to the lack of prior work on exactly the same task, we extend several state-of-the-art trajectory forecasting methods shown below as baselines. First, pedestrian detection results obtained in Section~\ref{subsec:datasets} were tracked over time with the SORT tracker~\cite{Bewley2016_sort} and interpolated linearly to create training and testing trajectory samples. Predicted future trajectories were mapped onto the input space and smoothed by a Gaussian filter to form future crowd density maps as done for the ground-truth maps in Section~\ref{subsec:setup}.
\begin{itemize}
    \item \textbf{Constant Velocity (ConstVel).} Despite the extensive work done to leverage deep recurrent models for trajectory forecasting, a simple approach that makes future trajectory predictions based on the velocity is considered a powerful approach. In accordance with \cite{scholler2019simpler}, we implemented a simple baseline that used the velocity at the last two frames of input sequences to linearly extrapolate future trajectories.
    \item \textbf{Social LSTM (S-LSTM)~\cite{Alahi_2016_CVPR}} is one of the most popular baselines used in many trajectory forecasting papers. It trains an LSTM network that forecasts trajectories for each individual while accepting hidden states of surrounding pedestrians as additional inputs. To train the S-LSTM, we adopted the hyper-parameter settings suggested in the paper.
    \item \textbf{Trajectron~\cite{Ivanovic_2019_ICCV}} is a state-of-the-art trajectory forecasting method that models pedestrian groups by a dynamic graph structure. Unlike S-LSTM, Trajectron can sample multiple future trajectories from a given deterministic input. In our experiment, we slightly modified the choices of hyper-parameters to better perform on our problem setting; specifically, we changed the considered distance between pedestrians to make it work on the image resolution of 640 $\times$ 480, and sampled 100 trajectories in the testing time.
\end{itemize}

\subsection{Results}
\paragraph{Comparisons with Baselines.}
Tables \ref{tbl:FDST_results} and \ref{tbl:ETH_results} show quantitative evaluations on the FDST and UCY datasets. We confirmed that PDFN outperformed baselines on all the datasets. DFN, which encoded crowd density maps into a single feature vector, showed degraded performances. These results demonstrate the advantage of using our patch-based forecasting approach. Baseline methods did not performed very well especially on the UCY where stable detection and tracking of pedestrians was not provided. Fig.~\ref{fig:result} visualizes typical results with several methods. While Trajectron could forecast future locations of several people accurately as shown in examples (a) and (b), it failed to capture many others that were not detected or not tracked in examples (c) and (d). In contrast, PDFN successfully forecast dynamics of individuals (examples (b) and (c)), a small group (examples (b) and (d)), and a crowd (example (a)).

\paragraph{Choices of Input Types.}
Overall, we found that PDFN-C, which accepted crowd density estimation results as input, obtained a higher performance with a lower standard deviation in terms of $D\sub{KL}$ (recall), and outperformed the other methods on the UCY dataset. These results demonstrate that the PDFN-C was able to cover diverse future patterns of crowd density dynamics. One reason is that the crowd density estimator used in our work~\cite{wang2019learning} gives a high recall performance in detecting crowds, which is necessary for our method to learn their diverse dynamics. Another possible reason is our choice of MSE loss to learn the patch-based dynamics in Eq.~(\ref{eq:mse}). When there are multiple possible future patterns of crowd density from similar inputs, the MSE loss would encourage the model to predict the mean of those future patterns. PDFN-D performed nicely under the metric of $D\sub{IKL}$ (precision) and worked comparably well against PDFN-C on the FDST dataset as shown in Table~\ref{tbl:FDST_results}.

\begin{table}[t]
\renewcommand{\arraystretch}{\arraystrechlen}
\center
\caption{\textbf{Effect of Gaussian kernel sizes}. (Average / Final) denotes the performance averaged over $T_{\mathrm{out}}$ output frames and performance at the final frame. All performances are averaged over the three videos in the UCY dataset.}
\label{tbl:kernel}
\adjustbox{max width=1.00\linewidth}{
\begin{tabular}{@{}lcccc@{}}
\toprule
  (Average/Final) & $\sigma$ & $D\sub{KL}$  & $D\sub{IKL}$  & $D\sub{JS}$  \\
\midrule
Trajectron \cite{Ivanovic_2019_ICCV} & 1 & 13.5 / 13.4& 9.67 / 11.1& 0.54 / 0.56\\
& 3 & 7.54 / 7.71& 3.90 / 4.37& 0.35 / 0.37 \\
& 6 & 3.69 / 3.83& 1.96 / 2.19& 0.23 / 0.24 \\
\midrule
\textbf{PDFN-C} & 1 & 5.98 / 10.4& 9.17 / 13.2& 0.49 / 0.60  \\
 & 3 & 0.89 / 1.44& 2.31 / 2.99& 0.18 / 0.25 \\
 & 6 & 0.39 / 0.54& 1.04 / 1.34& 0.09 / 0.13  \\
\bottomrule
\end{tabular}
}
\end{table}

\balance 

\paragraph{Effect of Gaussian Kernel Sizes.}
As discussed in Section~\ref{subsec:datasets}, we applied a Gaussian filter to input/output crowd density maps and ground truths to control with its kernel size $\sigma$ how strictly we evaluate the forecasted results. Table~\ref{tbl:kernel} shows how the averaged performance on the UCY dataset changed for different settings of $\sigma$. As expected, the overall performances of all methods degraded as the metrics became more strict with smaller $\sigma$. Even so, the proposed PDFN outperformed Trajectron.

\begin{figure}[t]
  \begin{center}
    \includegraphics[width=\linewidth]{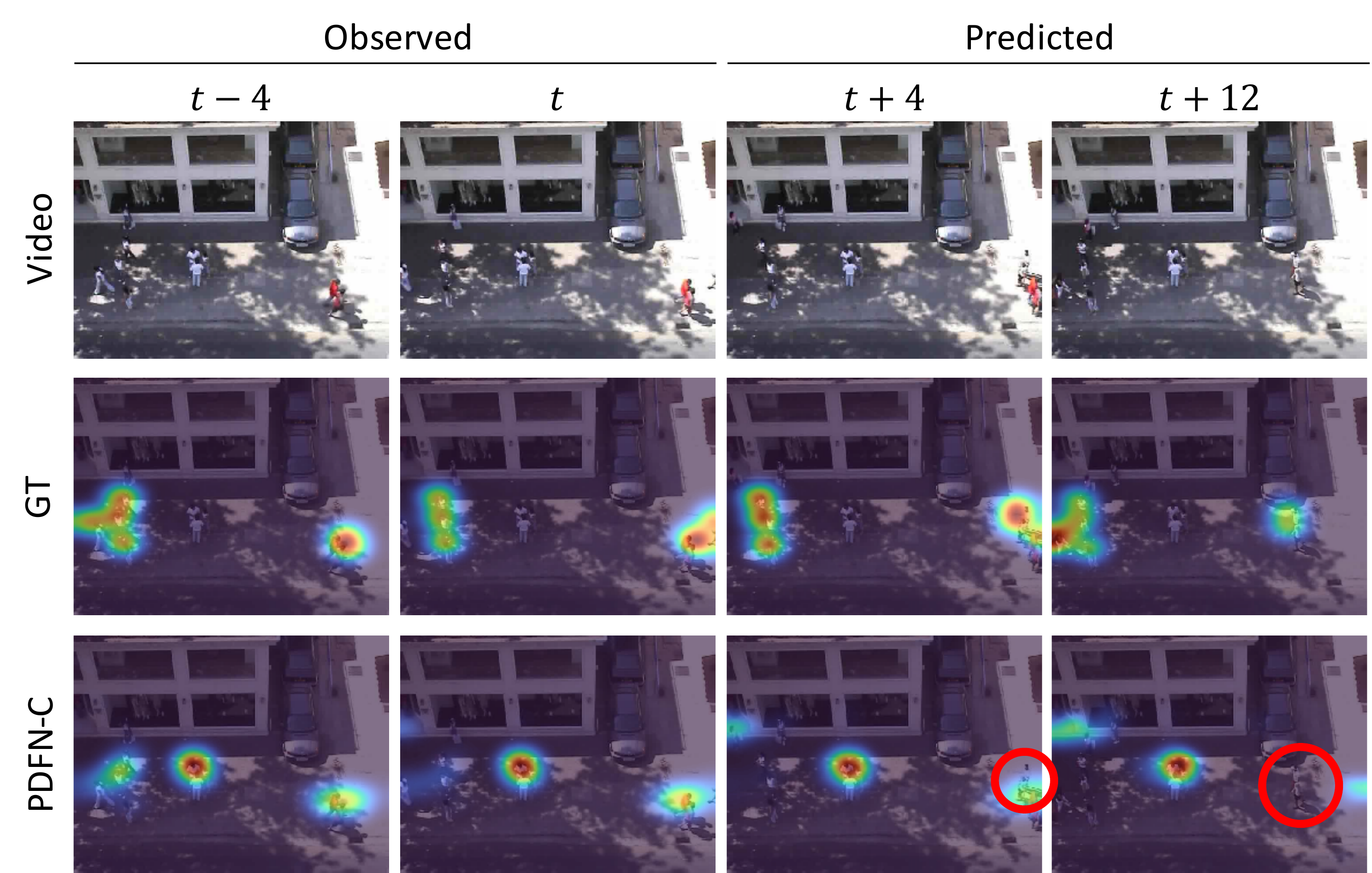}
    \caption{\textbf{Failure cases.} The presence of the pedestrian appearing from outside of the frame was not forecasted properly (highlighted in red circles).} 
    \label{fig:failure}
  \end{center}
\end{figure}

\paragraph{Failure Cases.}
One difficult case for all the methods including ours is the frame-in/out of a new person within future frames (such as the one highlighted in red circles in \ref{fig:failure}). This is generally an open challenge in visual forecasting and would require additional input sources such as videos recorded in surrounding locations~\cite{mazzon2012person} to enable person re-identification across multiple videos of a crowd.

\section{Conclusion}

This paper presented a new visual forecasting task named crowd density forecasting. By modeling patch-based dynamics, our approach was able to capture diverse and complex crowd density dynamics due to various numbers of independent groups of people. Moreover, contrary to existing trajectory forecasting approaches depending on accurate target detection and tracking, PDFN can utilize a high-performance crowd density estimation results as input to ensure accurate forecasting.

Extending crowd density forecasting approaches to work with not just a fixed surveillance camera but also wearable cameras~\cite{bertasius2018egocentric,manglik2019future,soo2016egocentric,rhinehart2019precog,yagi2018future} and auto-cameras~\cite{altche2017lstm,Chandra_2019_CVPR,houenou2013vehicle,Jain_2019_CoRL,Lee_2017_CVPR,Thiede_2019_ICCV,rhinehart2018r2p2,rhinehart2019precog} 
distributed throughout the world will lead to applications such as personalized navigation systems and intelligent driver assistance. Such applications would also require new computer vision techniques for re-identification of crowds across multiple cameras and long-term crowd dynamics modeling and forecasting.

\appendix

\section{Architecture Details}
Table~\ref{tab:model} below summarizes the specific architecture of PDFN used in our experiments. In the figure, Conv, Deconv, FC mean convolutional, deconvolutional, and fully-connected layers, respectively.

\begin{table}[h]
\begin{center}
\adjustbox{max width=1.00\linewidth}{
\begin{tabular}{@{}lcccc@{}}
\toprule
Layer type & \# channels & Kernel size & Stride size & Output size \\
\midrule
\multicolumn{5}{c}{\textbf{Patch Encoder Module}} \\
\midrule
Input & - & - & -& $80\times80\times1\times 8$ \\
2D-Conv+ReLU & 32 & $4 \times 4$ & $2 \times 2$ & $40\times40\times 32 \times8$ \\
2D-Conv+ReLU & 64 & $4 \times 4$ & $2 \times 2$ & $20\times20\times 64 \times8$ \\
2D-Conv+ReLU & 64 & $4 \times 4$ & $2 \times 2$ & $10\times10\times 64 \times8$ \\
FC & 16 & - & -& $10\times10\times16\times 8$ \\
\midrule
\multicolumn{5}{c}{\textbf{Patch-based Forecasting Module}} \\
\midrule
Transpose & - & - & - & $10\times10\times 8 \times 16 $  \\
3D-Conv+ReLU & 64 & $1 \times 1 \times 4$ & $1\times 1 \times 2$ & $10\times10\times 4 \times 64$ \\
3D-Conv+ReLU & 128 & $1 \times 1 \times 4$ & $1\times 1 \times 2$ & $10\times10\times 2 \times 128$ \\
3D-Conv+ReLU & 256 & $1 \times 1 \times 2$ & $1\times 1 \times 1$ & $10\times10\times 1 \times 256$ \\
3D-Deonv+ReLU & 128 & $1 \times 1 \times 3$ & $1\times 1 \times 1$ & $10\times10\times 3 \times 128$ \\
3D-Deonv+ReLU & 64 & $1 \times 1 \times 4$ & $1\times 1 \times 2$ & $10\times10\times 6 \times 64$ \\
3D-Deonv+ReLU & 16 & $1 \times 1 \times 4$ & $1\times 1 \times 2$ & $10\times10\times 12 \times 16$ \\
Transpose & - & - & - & $10\times10\times 16 \times 12 $  \\
\midrule
\multicolumn{5}{c}{\textbf{Patch Decoder Module}} \\
\midrule
2D-Deconv+ReLU & 32 & $4 \times 4$ & $2 \times 2$ & $20\times20\times32 \times 12$ \\
2D-Deconv+ReLU & 32 & $4 \times 4$ & $2 \times 2$ & $40\times40\times 32 \times 12$ \\
2D-Deconv+Sigmoid & 1 & $4 \times 4$ & $2 \times 2$ &  $80\times80\times 1\times 12$ \\
\bottomrule \\
\end{tabular}
}
\end{center}
\caption{{\bf PDFN architecture.}}
\label{tab:model}
\end{table}

\section{Base Performances of Crowd Density Maps}
In our experiments, we adopted two approaches to compute crowd density maps from video frames: crowd density estimation~\cite{wang2019learning} (used in PDFN-C) and pedestrian detection~\cite{lin2017feature} (in PDFN-D). One important question that supplements our main result is on the base performances given by these crowd density maps; in other words, \emph{``how much are the original crowd density maps similar to ground-truth maps?''} To address this question, we 1) extracted crowd density maps directly from future frames (\ie, $c_{t+1},\dots,c_{t+T\sub{out}}$) and 2) measured $D\sub{KL}, D\sub{IKL}, D\sub{JS}$ by using these maps as prediction results.

As shown in Tables~\ref{tbl:FDST_results_supp} and \ref{tbl:ETH_results_supp}, the original performances of crowd density maps were slightly better (\ie, lower divergences) than those of PDFN-C and PDFN-D in many cases. We also confirmed that results from the crowd density estimation outperformed those of the pedestrian detection in UCY. Note that this availability of crowd density estimation results for the input to forecasting models is a unique feature of the proposed patch-based approach, as it is not obvious how these results can be used to extract pedestrian trajectories for trajectory forecasting.

\begin{table*}[t]
\setlength{\tabcolsep}{14pt} % FIXME
\renewcommand{\arraystretch}{\arraystrechlen}
\center
\caption{\textbf{Base performances of crowd density maps on FDST}: $D\sub{KL}$, $D\sub{IKL}, D\sub{JS}$ for the original crowd density maps computed directly from future video frames (lower scores mean better fit to ground truth maps).} 
\label{tbl:FDST_results_supp}
\adjustbox{max width=1.00\linewidth}{
\begin{tabular}{@{}lcccccc@{}}
\toprule
& $D\sub{KL}$ (Average / Final)& $D\sub{IKL}$ (Average / Final) &$D\sub{JS}$ (Average / Final)\\
\midrule
Pedestrian Detection~\cite{lin2017feature} & 0.77 $\pm$ 1.21 / 0.94 $\pm$ 1.22& 0.61 $\pm$ 0.51 / 1.29 $\pm$ 1.01& 0.08 $\pm$ 0.06 / 0.12 $\pm$ 0.07\\
\textbf{PDFN-D} &0.75 $\pm$ 1.11 / 0.91 $\pm$ 1.12& 0.66 $\pm$ 0.55 / 1.30 $\pm$ 1.03 & 0.09 $\pm$ 0.06 / 0.13 $\pm$ 0.07 \\
\midrule
Crowd Density Estimation~\cite{wang2019learning}& 0.46 $\pm$ 0.42 / 0.61 $\pm$ 0.45& 1.47 $\pm$ 1.07 / 2.21 $\pm$ 1.61& 0.10 $\pm$ 0.06 / 0.14 $\pm$ 0.07\\
\textbf{PDFN-C} & 0.47 $\pm$ 0.41 / 0.63 $\pm$ 0.44 & 1.55 $\pm$ 1.10 / 2.25 $\pm$ 1.60& 0.10 $\pm$ 0.06 / 0.14 $\pm$ 0.07 \\
\bottomrule
\end{tabular}
}
\end{table*}

% TODO : Average / Final
\begin{table*}[t]
\setlength{\tabcolsep}{14pt} % FIXME
\renewcommand{\arraystretch}{\arraystrechlen}
\center
\caption{\textbf{Base performances of crowd density maps on UCY}: $D\sub{KL}$, $D\sub{IKL}, D\sub{JS}$ for the original crowd density maps computed directly from future video frames (lower scores mean better fit to ground truth maps).} 
\label{tbl:ETH_results_supp}
\adjustbox{max width=1.00\linewidth}{
\begin{tabular}{@{}lcccccc@{}}
\toprule
\textbf{Zara 1}& $D\sub{KL}$ (Average / Final)& $D\sub{IKL}$ (Average / Final) &$D\sub{JS}$ (Average / Final) \\
\midrule
Pedestrian Detection~\cite{lin2017feature} & 2.67 $\pm$ 1.98 / 3.35 $\pm$ 2.87& 5.35 $\pm$ 3.57 / 5.65 $\pm$ 4.17& 0.30 $\pm$ 0.11 / 0.32 $\pm$ 0.13\\
\textbf{PDFN-D} & 2.79 $\pm$ 1.98 / 3.81 $\pm$ 2.73& 5.48 $\pm$ 3.57 / 6.18 $\pm$ 4.07& 0.32 $\pm$ 0.11 / 0.38 $\pm$ 0.13\\
\midrule
Crowd Density Estimation~\cite{wang2019learning} & 0.70 $\pm$ 0.27 / 0.88 $\pm$ 0.61& 2.14 $\pm$ 1.65 / 2.34 $\pm$ 2.14& 0.15 $\pm$ 0.05 / 0.17 $\pm$ 0.07\\
\textbf{PDFN-C} & 0.87 $\pm$ 0.32 / 1.44 $\pm$ 0.69 & 2.28 $\pm$ 1.65 / 3.13 $\pm$ 2.19 & 0.18 $\pm$ 0.06 / 0.26 $\pm$ 0.09 \\
\bottomrule
\toprule
\textbf{Zara 2}& $D\sub{KL}$ (Average / Final)& $D\sub{IKL}$ (Average / Final) &$D\sub{JS}$ (Average / Final) \\
\midrule
Pedestrian Detection~\cite{lin2017feature} & 6.38 $\pm$ 4.03 / 7.00 $\pm$ 4.69& 6.83 $\pm$ 4.48 / 6.84 $\pm$ 4.91& 0.44 $\pm$ 0.12 / 0.45 $\pm$ 0.13\\
\textbf{PDFN-D}& 6.13 $\pm$ 4.00 / 6.46 $\pm$ 4.49& 6.82 $\pm$ 4.38 / 6.78 $\pm$ 4.68& 0.44 $\pm$ 0.12 / 0.44 $\pm$ 0.13\\
\midrule
Crowd Density Estimation~\cite{wang2019learning} & 0.73 $\pm$ 0.47 / 1.15 $\pm$ 1.00& 2.11 $\pm$ 2.58 / 2.47 $\pm$ 2.89& 0.15 $\pm$ 0.09 / 0.20 $\pm$ 0.10\\
\textbf{PDFN-C} & 0.86 $\pm$ 0.47 / 1.57 $\pm$ 1.02 & 2.29 $\pm$ 2.46 / 3.05 $\pm$ 2.94 & 0.17 $\pm$ 0.08 / 0.26 $\pm$ 0.10 \\
\bottomrule
\toprule
\textbf{University}& $D\sub{KL}$ (Average / Final)& $D\sub{IKL}$ (Average / Final) &$D\sub{JS}$ (Average / Final)\\
\midrule
Pedestrian Detection~\cite{lin2017feature} & 3.63 $\pm$ 1.38 / 4.26 $\pm$ 1.67& 3.20 $\pm$ 1.81 / 3.30 $\pm$ 1.97& 0.30 $\pm$ 0.06 / 0.32 $\pm$ 0.07\\
\textbf{PDFN-D} & 3.71 $\pm$ 1.38 / 4.54 $\pm$ 1.65& 3.31 $\pm$ 1.84 / 3.69 $\pm$ 2.10& 0.31 $\pm$ 0.06 / 0.35 $\pm$ 0.06\\
\midrule
Crowd Density Estimation~\cite{wang2019learning} & 0.98 $\pm$ 0.25 / 1.09 $\pm$ 0.33& 2.26 $\pm$ 0.84 / 2.33 $\pm$ 0.97& 0.19 $\pm$ 0.03 / 0.20 $\pm$ 0.04\\
\textbf{PDFN-C} & 0.99 $\pm$ 0.23 / 1.18 $\pm$ 0.32 & 2.42 $\pm$ 0.84 / 2.63 $\pm$ 0.99 & 0.20 $\pm$ 0.04 / 0.23 $\pm$ 0.05 \\
\bottomrule
\end{tabular}
}
\end{table*}

\balance
{\small
\bibliographystyle{ieee_fullname}

\input{egpaper_for_review.bbl}
}

\end{document}

%% file: result_table_fdst.tex
% TODO: ADE / FDE
\begin{table*}[t]
\setlength{\tabcolsep}{14pt} % FIXME
\renewcommand{\arraystretch}{\arraystrechlen}
\center
\caption{\textbf{Results on FDST}. The mean and standard deviation of $D\sub{KL}$, $D\sub{IKL}, D\sub{JS}$ were computed over 40 testing videos. (Average / Final) denotes the performance averaged over $T_{\mathrm{out}}$ output frames and performance at the final frame, respectively.} 
\label{tbl:FDST_results}
\adjustbox{max width=1.00\linewidth}{
\begin{tabular}{@{}lcccccc@{}}
\toprule
& $D\sub{KL}$ (Average / Final) & $D\sub{IKL}$ (Average / Final) &$D\sub{JS}$ (Average / Final)\\
\midrule
ConstVel \cite{scholler2019simpler}& 1.83 $\pm$ 2.15 / 2.30 $\pm$ 2.40& 0.92 $\pm$ 0.66 / 1.47 $\pm$ 1.19& 0.13 $\pm$ 0.08 / 0.18 $\pm$ 0.10 \\
S-LSTM \cite{Alahi_2016_CVPR} & 2.23 $\pm$ 2.33 / 2.80 $\pm$ 2.64& 1.89 $\pm$ 1.20 / 2.50 $\pm$ 1.98& 0.20 $\pm$ 0.09 / 0.24 $\pm$ 0.11 \\
Trajectron \cite{Ivanovic_2019_ICCV} & 1.60 $\pm$ 1.97 / 1.90 $\pm$ 2.13& 0.94 $\pm$ 0.92 / 1.59 $\pm$ 1.67& 0.12 $\pm$ 0.09 / 0.16 $\pm$ 0.10 \\
\midrule
DFN-D & 1.86 $\pm$ 0.69 / 1.88 $\pm$ 0.79& 5.35 $\pm$ 3.00 / 5.65 $\pm$ 2.99& 0.33 $\pm$ 0.07 / 0.33 $\pm$ 0.07 \\
DFN-C & 1.63 $\pm$ 0.51 / 1.68 $\pm$ 0.55& 6.67 $\pm$ 3.59 / 7.01 $\pm$ 3.68& 0.34 $\pm$ 0.08 / 0.34 $\pm$ 0.08\\
\textbf{PDFN-D} & 0.75 $\pm$ 1.11 / 0.91 $\pm$ 1.12& \textbf{0.66 $\pm$ 0.55 / 1.30 $\pm$ 1.03} & \textbf{0.09 $\pm$ 0.06 / 0.13 $\pm$ 0.07} \\
\textbf{PDFN-C}& \textbf{0.47 $\pm$ 0.41 / 0.63 $\pm$ 0.44} & 1.55 $\pm$ 1.10 / 2.25 $\pm$ 1.60& 0.10 $\pm$ 0.06 / 0.14 $\pm$ 0.07 \\
\bottomrule
\end{tabular}
}
\end{table*}

%% file: result_table_figure_ucy.tex
% TODO : Average / Final
\begin{table*}[t]
\setlength{\tabcolsep}{14pt} % FIXME
\renewcommand{\arraystretch}{\arraystrechlen}
\center
\caption{\textbf{Results on UCY}. The mean and standard deviation of $D\sub{KL}$, $D\sub{IKL}, D\sub{JS}$ were computed for each video sequence. (Average / Final) denotes the performance averaged over $T_{\mathrm{out}}$ output frames and performance at the final frame, respectively.} 
\label{tbl:ETH_results}
\adjustbox{max width=1.00\linewidth}{
\begin{tabular}{@{}lcccccc@{}}
\toprule
\textbf{Zara 1}& $D\sub{KL}$ (Average / Final)& $D\sub{IKL}$ (Average / Final) &$D\sub{JS}$ (Average / Final) \\
\midrule
ConstVel \cite{scholler2019simpler}& 7.60 $\pm$ 4.68 / 8.52 $\pm$ 5.06& 3.26 $\pm$ 3.53 / 4.52 $\pm$ 4.46& 0.33 $\pm$ 0.13 / 0.40 $\pm$ 0.15 \\
S-LSTM \cite{Alahi_2016_CVPR}& 10.9 $\pm$ 5.41 / 12.3 $\pm$ 5.47& 9.07 $\pm$ 5.97 / 10.5 $\pm$ 6.53& 0.50 $\pm$ 0.13 / 0.56 $\pm$ 0.13\\
Trajectron \cite{Ivanovic_2019_ICCV}& 7.19 $\pm$ 5.19 / 7.67 $\pm$ 5.56& 4.29 $\pm$ 4.81 / 5.36 $\pm$ 5.28& 0.34 $\pm$ 0.15 / 0.39 $\pm$ 0.15\\
\midrule
DFN-D& 4.82 $\pm$ 2.08 / 5.55 $\pm$ 2.77& 8.24 $\pm$ 4.21 / 9.31 $\pm$ 4.97& 0.48 $\pm$ 0.10 / 0.52 $\pm$ 0.12\\
DFN-C  & 0.93 $\pm$ 0.31 / 1.65 $\pm$ 0.73& 2.73 $\pm$ 1.63 / 4.06 $\pm$ 2.18& 0.20 $\pm$ 0.05 / 0.31 $\pm$ 0.08\\
\textbf{PDFN-D} & 2.79 $\pm$ 1.98 / 3.81 $\pm$ 2.73& 5.48 $\pm$ 3.57 / 6.18 $\pm$ 4.07& 0.32 $\pm$ 0.11 / 0.38 $\pm$ 0.13\\
\textbf{PDFN-C} & \textbf{0.87 $\pm$ 0.32 / 1.44 $\pm$ 0.69} & \textbf{2.28 $\pm$ 1.65 / 3.13 $\pm$ 2.19} & \textbf{0.18 $\pm$ 0.06 / 0.26 $\pm$ 0.09} \\
\bottomrule
\toprule
\textbf{Zara 2}& $D\sub{KL}$ (Average / Final)& $D\sub{IKL}$ (Average / Final) &$D\sub{JS}$ (Average / Final) \\
\midrule
ConstVel \cite{scholler2019simpler} & 7.76 $\pm$ 4.69 / 8.59 $\pm$ 4.85& 3.52 $\pm$ 3.93 / 4.41 $\pm$ 4.31& 0.34 $\pm$ 0.13 / 0.40 $\pm$ 0.13\\
S-LSTM \cite{Alahi_2016_CVPR} & 9.58 $\pm$ 4.78 / 10.8 $\pm$ 4.83& 7.48 $\pm$ 4.86 / 8.44 $\pm$ 5.56& 0.48 $\pm$ 0.12 / 0.52 $\pm$ 0.12\\
Trajectron \cite{Ivanovic_2019_ICCV} & 7.26 $\pm$ 4.77 / 7.41 $\pm$ 5.27& 3.69 $\pm$ 4.09 / 3.77 $\pm$ 4.49& 0.35 $\pm$ 0.13 / 0.34 $\pm$ 0.14\\
\midrule
DFN-D & 4.04 $\pm$ 2.18 / 3.59 $\pm$ 2.06& 7.53 $\pm$ 4.04 / 7.36 $\pm$ 4.23& 0.45 $\pm$ 0.11 / 0.44 $\pm$ 0.12\\
DFN-C & 0.99 $\pm$ 0.48 / 1.90 $\pm$ 1.03& 2.58 $\pm$ 2.48 / 3.70 $\pm$ 2.96& 0.20 $\pm$ 0.08 / 0.30 $\pm$ 0.10\\
\textbf{PDFN-D}& 6.13 $\pm$ 4.00 / 6.46 $\pm$ 4.49& 6.82 $\pm$ 4.38 / 6.78 $\pm$ 4.68& 0.44 $\pm$ 0.12 / 0.44 $\pm$ 0.13\\
\textbf{PDFN-C} & \textbf{0.86 $\pm$ 0.47 / 1.57 $\pm$ 1.02} & \textbf{2.29 $\pm$ 2.46 / 3.05 $\pm$ 2.94} & \textbf{0.17 $\pm$ 0.08 / 0.26 $\pm$ 0.10} \\
\bottomrule
\toprule
\textbf{University}& $D\sub{KL}$ (Average / Final)& $D\sub{IKL}$ (Average / Final) &$D\sub{JS}$ (Average / Final)\\
\midrule
ConstVel \cite{scholler2019simpler} & 8.92 $\pm$ 3.23 / 8.99 $\pm$ 3.22& 3.86 $\pm$ 2.30 / 4.13 $\pm$ 2.33& 0.39 $\pm$ 0.08 / 0.41 $\pm$ 0.08\\
S-LSTM \cite{Alahi_2016_CVPR}& 8.96 $\pm$ 2.98 / 9.70 $\pm$ 2.82& 6.21 $\pm$ 3.02 / 6.99 $\pm$ 3.52& 0.47 $\pm$ 0.07 / 0.52 $\pm$ 0.07\\
Trajectron \cite{Ivanovic_2019_ICCV} & 8.63 $\pm$ 3.47 / 8.34 $\pm$ 3.47& 3.69 $\pm$ 2.31 / 3.93 $\pm$ 2.33& 0.37 $\pm$ 0.08 / 0.39 $\pm$ 0.08\\
\midrule
DFN-D & 9.07 $\pm$ 1.89 / 8.82 $\pm$ 2.12& 4.90 $\pm$ 2.15 / 5.13 $\pm$ 2.88& 0.48 $\pm$ 0.06 / 0.48 $\pm$ 0.07\\
DFN-C& 1.04 $\pm$ 0.25 / 1.35 $\pm$ 0.35& 2.49 $\pm$ 0.91 / 2.98 $\pm$ 1.09& 0.21 $\pm$ 0.04 / 0.26 $\pm$ 0.05\\
\textbf{PDFN-D} & 3.71 $\pm$ 1.38 / 4.54 $\pm$ 1.65& 3.31 $\pm$ 1.84 / 3.69 $\pm$ 2.10& 0.31 $\pm$ 0.06 / 0.35 $\pm$ 0.06\\
\textbf{PDFN-C} & \textbf{0.99 $\pm$ 0.23 / 1.18 $\pm$ 0.32} & \textbf{2.42 $\pm$ 0.84 / 2.63 $\pm$ 0.99} & \textbf{0.20 $\pm$ 0.04 / 0.23 $\pm$ 0.05} \\
\bottomrule
\end{tabular}
}
\end{table*}

%% file: result_figure.tex
\def\figcolwidth{0.47}
\begin{figure*}[t]
 \centering
  \begin{minipage}[t]{\figcolwidth\linewidth}
    \centering
    \includegraphics[width=1.0\linewidth]{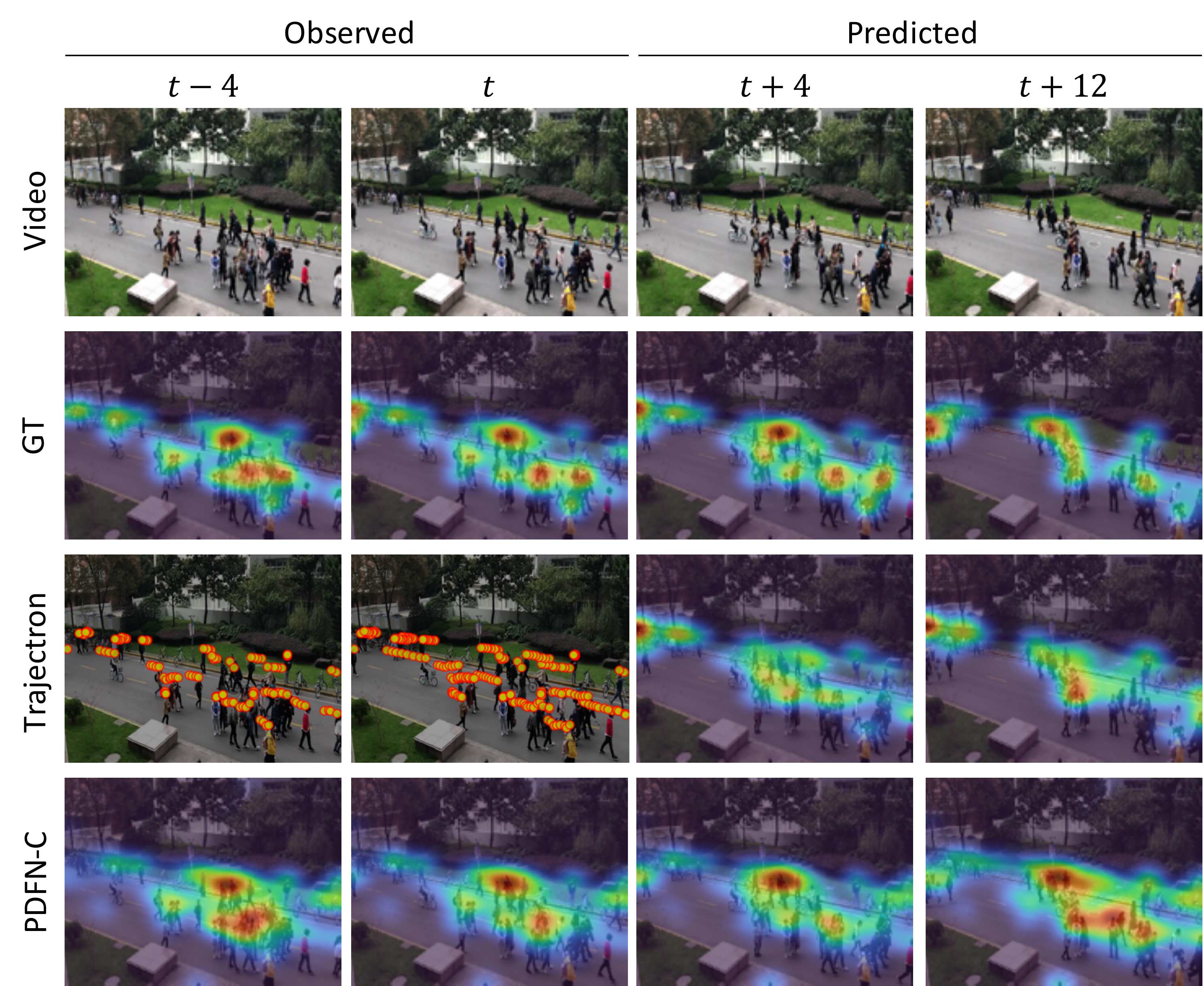} 
    \subcaption{}
  \end{minipage} 
  \begin{minipage}[t]{\figcolwidth\linewidth}
    \centering
    \includegraphics[width=1.0\linewidth]{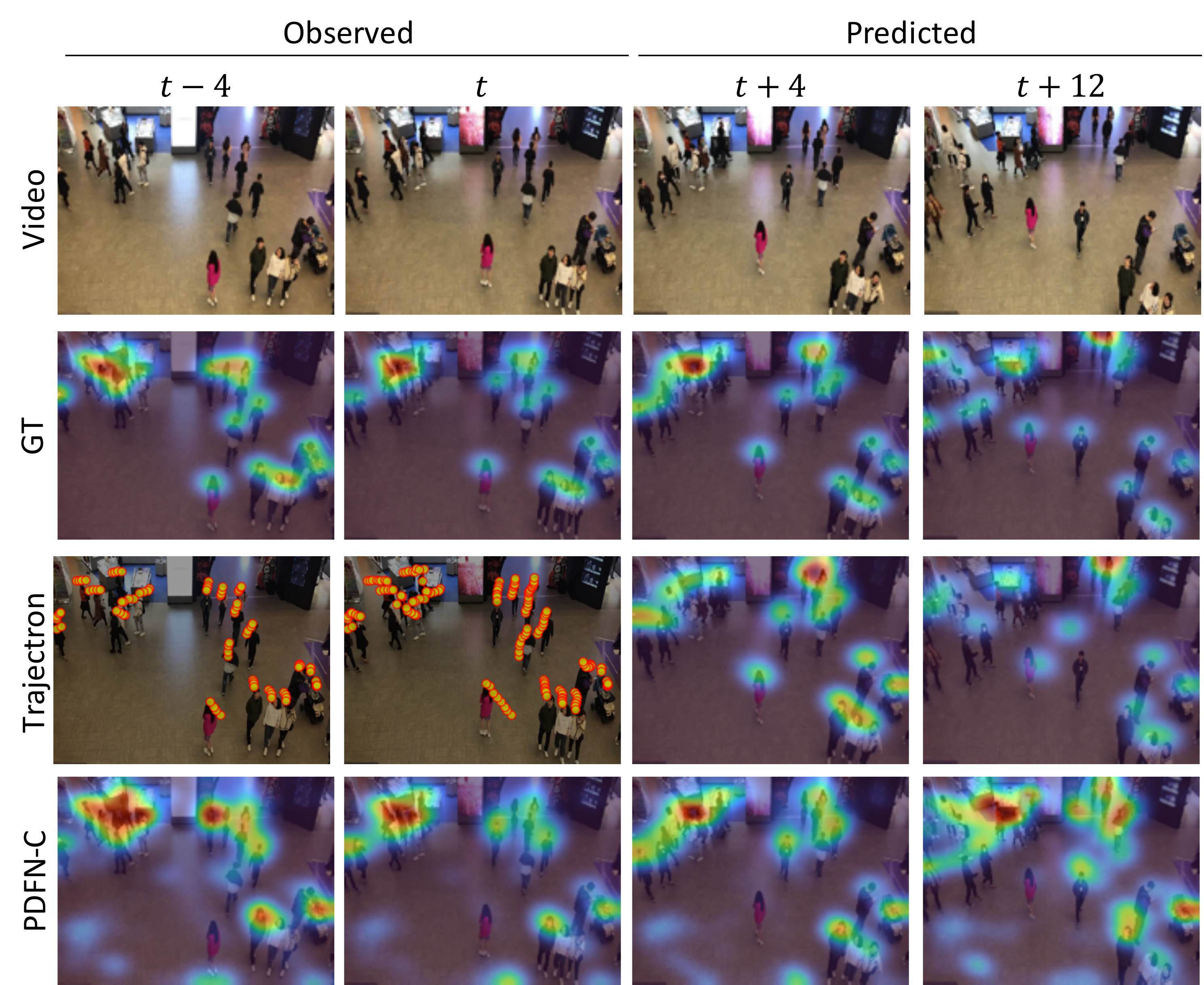} 
    \subcaption{}
  \end{minipage} 
  \\
  \begin{minipage}[t]{\figcolwidth\linewidth}
    \centering
    \includegraphics[width=1.0\linewidth]{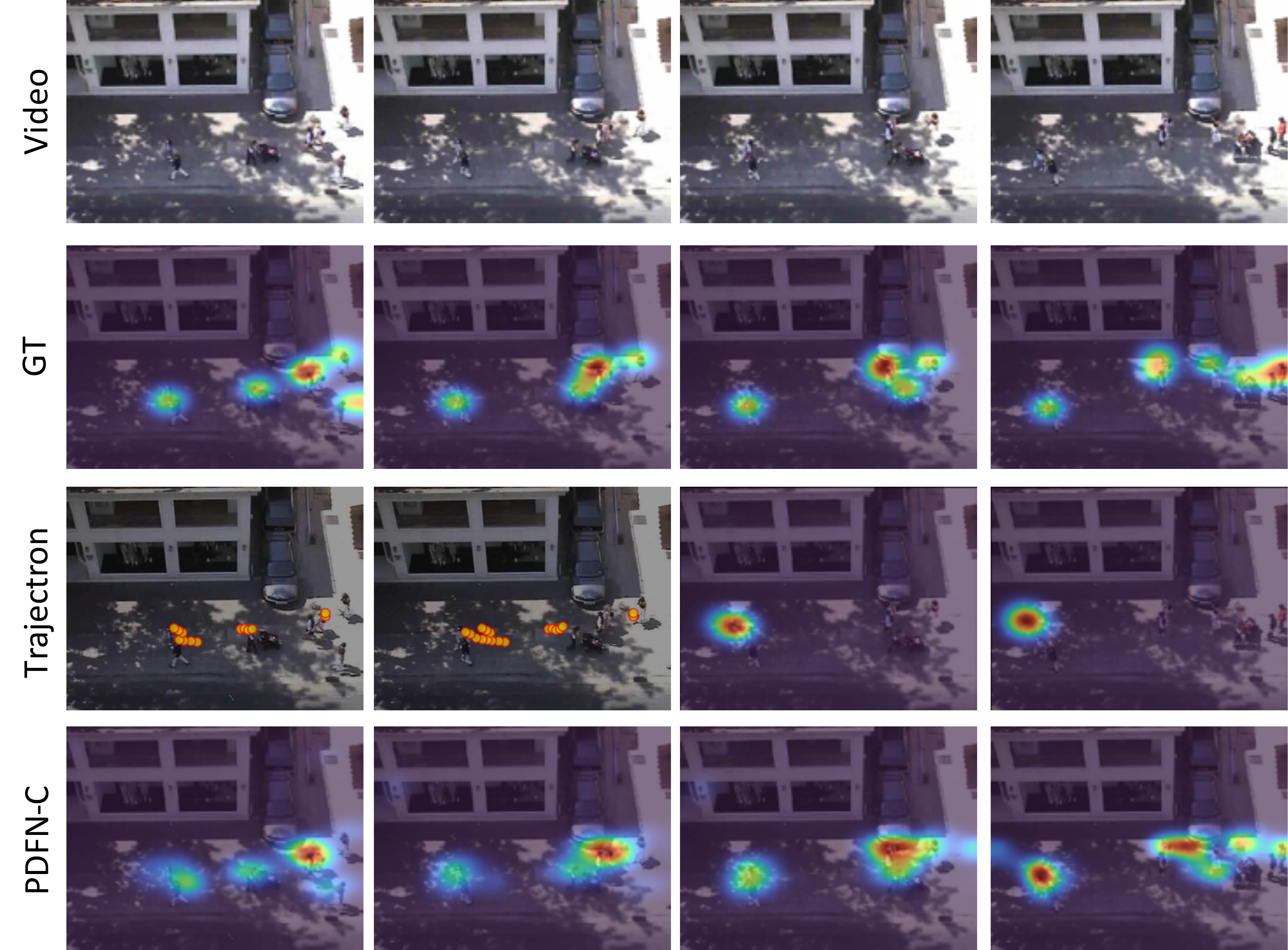} 
    \subcaption{}
 \end{minipage}%%
 \begin{minipage}[t]{\figcolwidth\linewidth}
    \centering
    \includegraphics[width=1.0\linewidth]{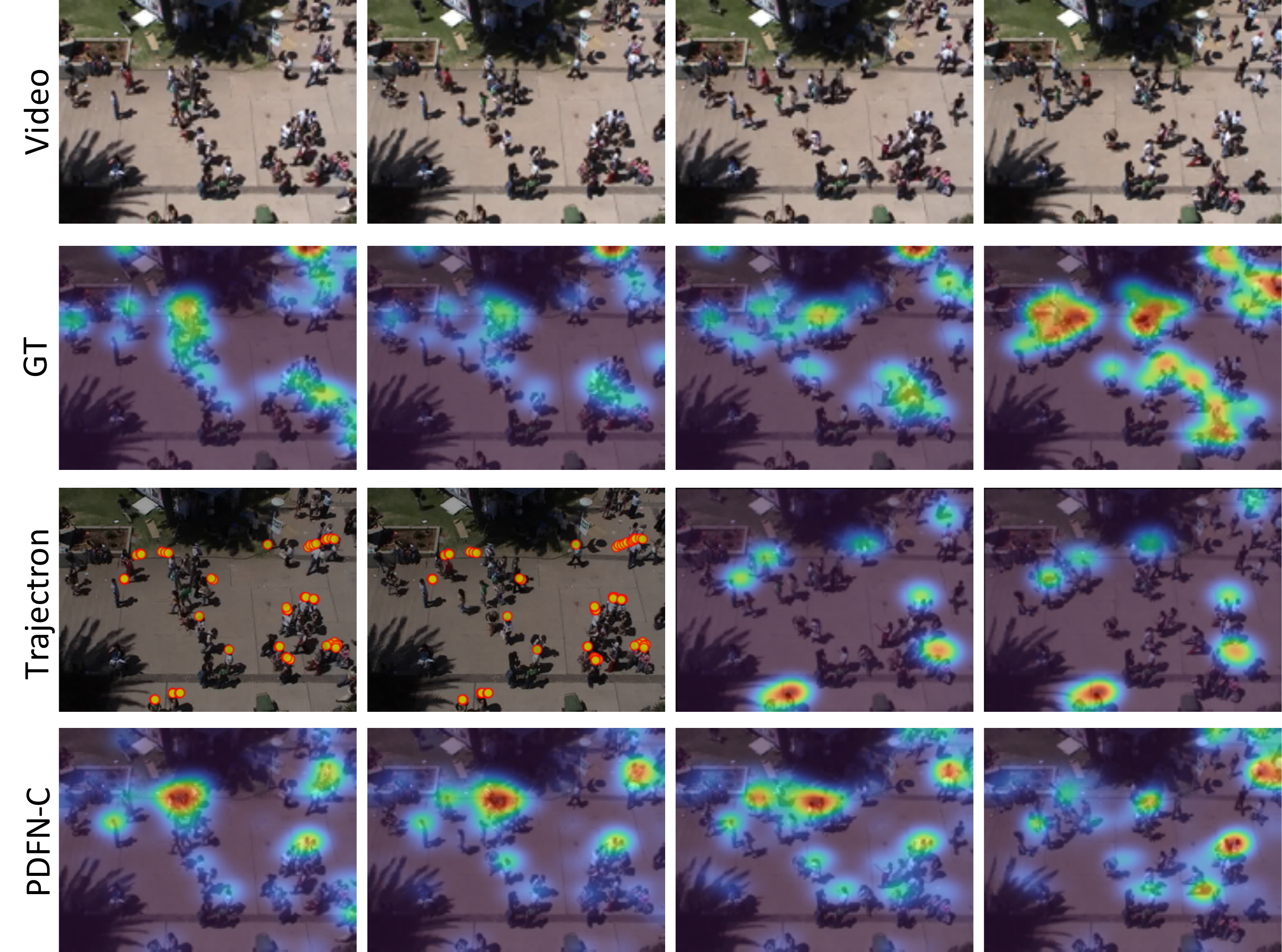} 
    \subcaption{}
  \end{minipage} 
 \caption{\textbf{Qualitative results of selected methods.} Input, ground-truth (GT), and predicted crowd density maps are laid over original video frames. Future video frames after time $t$ are not available in the evaluation phase, and the corresponding crowd density maps are predicted from their history extracted from past video frames. Detected and tracked pedestrians for Trajectron are described in yellow circles.}
 \label{fig:result}
\end{figure*}